\documentclass{article}
\usepackage{ijcai19}
\usepackage[none]{hyphenat}
\usepackage{microtype}
\usepackage{times}
\usepackage{soul}
\usepackage{url}
\usepackage[hidelinks]{hyperref}
\usepackage[utf8]{inputenc}
\usepackage[small]{caption}
\usepackage{graphicx}
\usepackage{amsmath}
\usepackage{booktabs}
\usepackage{bm}

\usepackage[utf8]{inputenc}
\usepackage[english]{babel}
\usepackage{amsthm}
\usepackage{amssymb}
\usepackage{amsmath}

\newtheorem{theorem}{Theorem}
\newtheorem{corollary}{Corollary}[theorem]
\newtheorem{lemma}{Lemma}[theorem]
\usepackage{algorithm}
\usepackage[noend]{algpseudocode}

\usepackage{hhline}
\usepackage{multirow}
\usepackage{array}

\makeatletter
\def\BState{\State\hskip-\ALG@thistlm}
\makeatother

\newcolumntype{P}[1]{>{\centering\arraybackslash}p{#1}}
\newcolumntype{M}[1]{>{\centering\arraybackslash}m{#1}}

\emergencystretch=\maxdimen
\title{Compressibility Loss for Neural Network Weights}

\author{
Caglar Aytekin, Francesco Cricri and Emre Aksu
\affiliations
Nokia Technologies
}

\begin{document}
\begin{sloppypar}
\maketitle

\begin{abstract}
  In this paper we apply a compressibility loss that enables learning highly compressible neural network weights. The loss was previously proposed as a measure of negated sparsity of a signal, yet in this paper we show that minimizing this loss also enforces the non-zero parts of the signal to have very low entropy, thus making the entire signal more compressible. For an optimization problem where the goal is to minimize the compressibility loss (the objective), we prove that at any critical point of the objective, the weight vector is a ternary signal and the corresponding value of the objective is the squared root of the number of non-zero elements in the signal, thus directly related to sparsity. In the experiments, we train neural networks with the compressibility loss and we show that the proposed method achieves weight sparsity and compression ratios comparable with the state-of-the-art.
\end{abstract}

\section{Introduction}
In recent years, neural networks have shown outstanding performance in several tasks, at the expense of a high number of model parameters or weights \cite{Simonyan2014}, \cite{He2016}, \cite{Szegedy2017}. As the initial main goal was to maximize performance, such as classification accuracy, the model size was not considered to be an issue, especially given the significant computational and memory resources of typical servers used for training and inference. However, alternative platforms such as mobile devices and edge-based embedded devices come with limited computational and memory capabilities, thus imposing strict limitations on the size of neural networks. 

In order to reduce either the storage requirements of neural networks, or their inference complexity, or both, several algorithms have addressed the reduction of parameters, while retaining most of the original model's performance. This is possible as neural networks seem to contain a significant amount of redundancy in their weights. 

One popular approach aims at pruning weights and convolution filters which are considered as least important based on their contribution, with a successive fine-tuning phase which adapts the network to the architectural changes introduced by the pruning operation. In \cite{Lecun1989}, the contribution of each weight is determined as its effect to the training error when setting its value to zero. 

Quantization approaches focus on reducing the precision or bit-depth of weights, as in \cite{Hubara2016}. For example, the authors in \cite{Vanhoucke2019} showed that quantizing weights to 8 bits results in significant speed-up while incurring into minimal accuracy losses. 

The work in \cite{Han2016} combines different approaches, namely pruning, quantization and Huffman-coding. Regarding pruning, the contribution of a weight is determined by its absolute numerical value, thus weights with low absolute values with respect to a threshold are pruned. In addition, the authors quantize the non-pruned weights by clustering them into a set of clusters using k-means (256 clusters for convolutions and 32 clusters for fully-connected layers).

Yet another approach is the low-rank factorization of weight matrices, where a matrix is factorized or decomposed into two lower-rank matrices. In \cite{Chen2018}, the authors note that a small rank can limit the expressiveness of the model and thus propose to learn an input-dependent factorization. 

Here, we propose to use a loss that was originally proposed in \cite{Hoyer} to achieve sparsity in a signal. This allows for training a neural network to have as many weights with zero value as possible. Moreover, in this paper we show that at the same time the loss also approximates a quantization operation which maps values of non-zero weights to a discrete number of possible values. 

The paper is organized as follows. Section \ref{related_work} surveys previous works which are most closely related to the method proposed in this paper. Section \ref{proposed_method} describes the proposed method in detail. In particular, it introduces the compressibility loss, its properties, and the post-training quantization strategy. Section \ref{results} describes our experiments and results, provides in-depth analyses of the effects of sparsity and quantization, and reports on the application of curriculum-learning to the compressibility loss. Section \ref{conclusions} discusses the obtained results and future directions. 

\section{Related Work}
\label{related_work}
There have been several previous works on sparsifying weights in neural networks by using a sparsity constraint during training. In \cite{Lebedev2016}, the authors use group-sparsity regularization on convolutional filters, which has the effect of shrinking some groups to zero. In particular, they use a $l_{2,1}$-norm regularizer. In \cite{Zhou2016} the authors state that in FFT domain (used to accelerate matrix multiplications on GPUs) sparse matrices are not anymore sparse, thus they propose a method for pruning neurons instead of simply weights. To this end, they impose a sparsity constraint during training directly on the neurons, for which two options are experimented with: tensor low rank constraint and group sparsity constraint (also here a $l_{2,1}$-norm regularizer). 

In \cite{Gomez2018}, the authors propose a targeted dropout where the weights of the neural network are first sorted at descending order according to their absolute values and the bottom $\gamma$ portion of weigts are randomly (with probability $\alpha$) and independently disabled/dropped during training. The work aims only sparsity and does not consider the compressibility of the remaining weights.

In \cite{Leclerc2018}, the authors propose a method to train a neural network with $L_1$ loss on a switch variable which takes a value between 0 and 1 and is used to be multiplied the weight value. During training, the weights that are very small are actually pruned and network adaptively changes architecture.

In \cite{Choi2018}, a universal neural network compression method was proposed that applies uniform random dithering and lattice quantization on neural network weights followed by a fine-tuning process.

\section{Proposed Method}
\label{proposed_method}
\subsection{Compressibility Loss}
The neural network compression method is based on the following compressibility loss which was partially introduced in \cite{Hoyer}:

\begin{equation}
    L(\bm{x})=\frac{||\bm{x}||_1}{||\bm{x}||_2}
    \label{loss}
\end{equation}

where $||\bm{x}||_1$ and $||\bm{x}||_2$ stand for $L_1$ and $L_2$ norms of $\bm{x}$ respectively. Originally this loss was proposed as a negated measure of sparsity.

Next, we show via following theorem that besides sparsity, the loss also helps achieving a very low entropy for the non-zero part of the signal as well.

\begin{theorem}
Let $\bm{x} \in \mathbb{R}^d$ be any non-zero vector, then at any critical point of $L(\bm{x})$, $\bm{x}$ is a ternary signal which satisfies $x_i \in \{ -c,0,c \}$ where $c=\frac{(||\bm{x}||_2)^2}{||\bm{x}||_1}$.
\label{main}
\end{theorem}

\begin{proof}
Let us begin by taking the derivative of $L(\bm{x})$ with respect to $\bm{x}$ as formulated in Eq. \ref{deriv}.
\begin{equation}
    \frac{\partial L(\bm{x})}{\partial \bm{x}} = \frac{sign(\bm{x})}{||\bm{x}||_2}-\frac{\bm{x}||\bm{x}||_1}{(||\bm{x}||_2)^3}
    \label{deriv}
\end{equation}

Note that in Eq. \ref{deriv}, $sign(.)$ function applies element-wise. Then, we equate the derivative to zero to find the critical points which gives the equality in Eq. \ref{derivzero}.


\begin{equation}
	\bm{x} = \frac{sign(\bm{x}) (||\bm{x}||_2)^2}{||\bm{x}||_1}
    \label{derivzero}
\end{equation}

We analyse Eq. \ref{derivzero} by considering the following three cases.

\begin{itemize}
  \item $x_i>0$ 
  In this case Eq. \ref{derivzero} reduces to:
  \begin{equation}
    x_i=\frac{(||\bm{x}||_2)^2}{||\bm{x}||_1}
    \label{case1}
  \end{equation}
  \item $x_i<0$ 
  In this case Eq. \ref{derivzero} reduces to:
  \begin{equation}
    x_i=-\frac{(||\bm{x}||_2)^2}{||\bm{x}||_1}
    \label{case2}
  \end{equation}
  \item $x_i=0$ 
  In this case Eq. \ref{derivzero} reduces to:
  \begin{equation}
    x_i=0
    \label{case3}
  \end{equation}
\end{itemize}

The Equations \ref{case1}, \ref{case2} and \ref{case3} prove Theorem \ref{main}.

\end{proof}

\begin{corollary}
The objective $L(\bm{x})$ at a critical point takes the value of $\sqrt{n}$ where $n$ is the number of nonzero elements in $\bm{x}$.
\label{maincorol}
\end{corollary}

\begin{proof}
According to Theorem \ref{main}, at a critical point, $x_i \in \{ -c,0,c \}$ where $c=\frac{(||\bm{x}||_2)^2}{||\bm{x}||_1}$ holds, therefore the following also holds.

\begin{equation}
    ||\bm{x}||_1=n\frac{(||\bm{x}||_2)^2}{||\bm{x}||_1}
    \label{corolproof1}
\end{equation}

Then, Eq. \ref{corolproof1} can be re-arranged as follows.

\begin{equation}
    \sqrt{n}=\frac{||\bm{x}||_1}{||\bm{x}||_2}
    \label{corolproof2}
\end{equation}

Eq. \ref{corolproof2} proves Corollary \ref{maincorol}.
 
\end{proof}

In order to make use of the loss in Eq. \ref{loss} and the result of Theorem \ref{main}, one has to prove that the critical points are in fact local minima as it is stated in Theorem \ref{local}.

\begin{theorem}
All critical points $\bm{x}$ of $L(\bm{x})$ are local minima from the directions where the direction vector is non-zero in at least one point which satisfies $x_i=0$.
\label{local}
\end{theorem}

The proof for Theorem \ref{local} is provided in the Appendix \ref{proofsec}.

\subsection{Properties of Compressibility Loss}

The compressibility loss differentiates from the commonly used $L_2$ and $L_1$ regularizers, which shrink the weights and help to achieve sparsification. By shrinking the weights, these regularizers do not obtain more compressible nonzero weights. 
The compressibility loss not only sparsifies the weights, but also ensures the non-zero elements are highly compressible by enforcing the signal to be ternary at critical points. 
In addition, as can be seen from Theorem \ref{main}, at a critical point, the non-zero part of the weight vector ($\pm{c}$) can take any value.

Moreover, compressibility loss is a better indicator of sparsity. At the same level of sparsity, independent of the values of nonzeros, the compressibility loss gives the same value at critical points. On the contrary, for  $L_2$ and $L_1$ losses this is not ensured.

\subsection{Compression Method}
\label{compmet}

We propose a neural network compression method which consists of retraining a given network architecture using the compressibility loss. 
A straightforward way to train a network to be compressible would be to apply the compressibility loss separately to each weight layer $\bm{w}^{(i)}, i \in [1,l]$ in the neural network. The compressibility loss of this method can be formulated as in Eq. \ref{alt1}. 

\begin{equation}
L_{c_1}= \sum_{i=1}^l \lambda_i L(\bm{w}^{(i)})
\label{alt1}
\end{equation}

\begin{table*}
  \centering
  \renewcommand{\arraystretch}{1.2}
  \begin{tabular}{|P{1.3cm}|P{0.6cm}|P{0.8cm}|P{0.6cm}|P{0.8cm}|P{0.6cm}|P{0.8cm}|P{0.6cm}|P{0.8cm}|P{0.6cm}|P{0.8cm}|P{0.6cm}|P{0.8cm}|P{0.6cm}|P{0.8cm}|P{0.6cm}|P{0.8cm}|P{0.6cm}|P{0.8cm}}
    \hline
    \multirow{2}{1.6cm}{\textbf{Sparsity}}  & \multicolumn{2}{c|}{\textbf{Original}} & \multicolumn{2}{c|}{\textbf{CNET$_{\text{0.005}}$}} & \multicolumn{2}{c|}{\textbf{CNET$_{\text{0.01}}$}} & \multicolumn{2}{c|}{\textbf{CNET$_{\text{0.02}}$}} & \multicolumn{2}{c|}{\textbf{CNET$_{\text{0.03}}$}} & \multicolumn{2}{c|}{\textbf{CNET$_{\text{0.045}}$}}\\
    \cline{2-13}
     & \textbf{Acc.} & \textbf{Comp.} & \textbf{Acc.} & \textbf{Comp.} & \textbf{Acc.} & \textbf{Comp.} & \textbf{Acc.} & \textbf{Comp.} & \textbf{Acc.} & \textbf{Comp.} & \textbf{Acc.} & \textbf{Comp.}\\
    \hline
    0$\%$ & 91.51 & 1 & 91.36 & 1 & 90.97 & 1 & 90.30 & 1 & 89.46 & 1 & 88.41 & 1  \\ \hline
    30$\%$  & 88.96 & 1.42 & 91.36 & 1.36 & 90.97 &  1.36 & 90.30 & 1.35 & 89.46 & 1.36 & 88.41 & 1.37  \\ \hline
    40$\%$ &  83.72 & 1.66 & 91.36 & 1.61 & 90.97 & 1.56 & 90.30 & 1.57 & 89.46 & 1.58 & 88.41 & 1.59 \\ \hline
    50$\%$ &  61.76 & 1.98 & 91.26 & 1.94 & 90.97 & 1.88 & 90.30 & 1.86 & 89.46 & 1.88 & 88.41 & 1.88 \\ \hline
    60$\%$ &  30.68 & 2.45 & 91.02 & 2.43 & 90.98 & 2.41 & 90.30 & 2.31 & 89.46 & 2.33 & 88.41 & 2.32  \\ \hline
    70$\%$ & 10.04 & 3.23 & 89.37 & 3.24 & 90.71 & 3.22 & 90.30 & 3.16 & 89.46 & 3.10 & 88.41 & 3.06  \\ \hline
    80$\%$ & 10.00 & 4.88 & 67.18 & 4.84 & 86.83 & 4.73 & 90.02 & 4.75 & 89.46 & 4.77 & 88.41 & 4.61  \\ \hline
    90$\%$ & 10.00 & 9.33 & 12.29 & 9.18 & 33.10 & 9.24 & 82.68 & 9.30 & 84.85 & 9.32 & 88.09 & 9.31   \\ \hline
  \end{tabular}
  \caption{Sparsity Effect: Accuracy (Acc.) and Mask Compression Ratio (Comp.) comparison to baseline at different sparsity levels.}
  \label{basecomp1}
\end{table*}

This method would require manually specifying the $\lambda_i$ for the expected compressibility at each layer of the neural network. 
A simple implementation can be to achieve the same level of compressibility in each layer by setting all $\lambda_i$ to the same value.
However this would assume that each layer has the same amount of redundancy which is not necessarily true.
Another drawback of this method is that non-zero elements at critical points may take different values in different layers, which is not desirable from a compressibility point of view. 
For example, let us assume an ideal case where two weight vectors in two layers ended up in a critical point of compressibility loss, i.e. $w_i^{(1)} \in \{-c_0,0,c_0\}$ and $w_i^{(2)} \in \{-c_1,0,c_1\}$. 
Although both vectors are ternary, the concatenation of these vectors is not necessarily ternary, i.e. $c_0=c_1$ does not necessarily hold. 
This limits the compressibility to a layer-wise level.

Because of the limitations of the above method, we propose to apply the compressibility loss on a single vector $\bm{w_{net}}$ obtained by flattening and concatenating every weight tensor in the neural network. The compressibility loss of this method can be formulated as in Eq. \ref{alt2}.

\begin{equation}
L_{c_2}= \lambda L(\bm{w_{net}})
\label{alt2}
\end{equation}

Applying the loss on a single concatenated weight vector would ensure compressibility at neural network level as opposed to layer-wise level in the first method.
For example, in the ideal case of ending up at a local minimum of compressibility loss, one can guarantee that any weight in the neural network would satisfy $w_{net_i} \in \{-c,0,c\}$, hence the entire neural network's weight vector is ternary.
Moreover, when using the compressibility loss in combination with a task-specific loss (such as categorical cross-entropy loss for image classification), the sparsity level at each layer would be adjusted automatically according to task performance. I.e., the sparsity level of each layer would be proportional to the redundancy of that layer with respect to the final task, thus avoiding making assumptions on the redundancy of each layer.

Based on the discussed advantages throughout the paper we only use the compressibility loss applied to entire network as in Eq. \ref{alt2}.

\subsubsection{Post-Training Pruning and Non-uniform Quantization}
After the training, we prune the neural network weights by simply setting the weights smaller than a threshold to zero. 
After pruning, we perform a non-uniform quantization based on $k$-means clustering. This is applied to the non-pruned weights of the entire neural network.
The number of clusters $k$ is selected based on the desired quantization level.

\subsubsection{Coding and Compression}

Instead of including the pruned weights (zero values) in the encoded weight vector, we create a separate binary mask which indicates zeros and non-zeros in the weight vector.
The mask is then flattened, bit-packed and compressed using Numpy npz algorithm \cite{npz2019}.
Then, the $k$-means labels of weight elements are stored in another tensor and npz-compressed.
Finally the cluster centroids for $k$-means are stored and npz-compressed.
The compression ratio is calculated as the ratio of the total file size of the above files over the file size of the npz-compressed original neural network weights.

\section{Experimental Results}
\label{results}

We conduct experiments on the image classification task using ResNet32 architecture.
The baseline method consists of training on CIFAR-10 dataset using categorical cross-entropy loss.
Our method adds the compressibility loss (given in Eq. \ref{alt2}) during training.
We conduct trainings with different $\lambda$ settings for compressibility loss and report the accuracy and compression levels at different sparsity levels.
A sparsity level is simply achieved by pruning the weights by the threshold that achieves the desired sparsity level.
We refer to our method as compressible network (CNET) and we indicate the $\lambda$ setting as a subscript as $CNET_{\lambda}$.

\subsection{Sparsity Effect}

In order to check our method's effect on resulting weight sparsity, we first report the accuracies and compression ratios without post-training non-uniform quantization method described in Section \ref{compmet}. 
Therefore the reported accuracies in Table \ref{basecomp1} are not affected by the quantization, but only pruning.

Moreover, the reported compression ratio only evaluates the effect due to mask-coding of zeros. In this experiment, non-zero values are not quantized at all but only npz compressed.
As observed from the Table \ref{basecomp1}, as we increase $\lambda$, the accuracy without pruning decreases gradually, yet the accuracy becomes more robust to pruning.
For example with $\lambda=0.005$, the accuracy without pruning is almost the same with original network, however after sparsity level $60\%$, the accuracy drops steadily. 
With $\lambda=0.045$, the accuracy without pruning is $3\%$ lower than the original network, however the network is robust to pruning and there is almost no accuracy drop until sparsity level $90\%$.
The compression ratios are similar for all models at same sparsity levels. 
This is expected since the only compression is due to mask-coding the zeros and non-zeros are not compressed. 
The only difference in compression ratios may be resulting from the very small deviation in sparsity levels.

\subsection{Quantization Effect}

The second experiment highlights the compressibility of non-zero weights aspect of our method.
Compression of non-zeros is achieved by the method described in Section \ref{compmet}.
At sparsity level $0\%$ we do not perform compression and report the accuracy directly.
At other sparsity levels, we always apply the compression method described in Section \ref{compmet}. 

\begin{table*}
  \centering
  \renewcommand{\arraystretch}{1.2}
  \begin{tabular}{|P{1.3cm}|P{0.6cm}|P{0.8cm}|P{0.6cm}|P{0.8cm}|P{0.6cm}|P{0.8cm}|P{0.6cm}|P{0.8cm}|P{0.6cm}|P{0.8cm}|P{0.6cm}|P{0.8cm}|P{0.6cm}|P{0.8cm}|P{0.6cm}|P{0.8cm}|P{0.6cm}|P{0.8cm}}
    \hline
    \multirow{2}{1.6cm}{\textbf{Sparsity}}  & \multicolumn{2}{c|}{\textbf{Original}} & \multicolumn{2}{c|}{\textbf{CNET$_{\text{0.005}}$}} & \multicolumn{2}{c|}{\textbf{CNET$_{\text{0.01}}$}} & \multicolumn{2}{c|}{\textbf{CNET$_{\text{0.02}}$}} & \multicolumn{2}{c|}{\textbf{CNET$_{\text{0.03}}$}} & \multicolumn{2}{c|}{\textbf{CNET$_{\text{0.045}}$}}\\
    \cline{2-13}
     & \textbf{Acc.} & \textbf{Comp.} & \textbf{Acc.} & \textbf{Comp.} & \textbf{Acc.} & \textbf{Comp.} & \textbf{Acc.} & \textbf{Comp.} & \textbf{Acc.} & \textbf{Comp.} & \textbf{Acc.} & \textbf{Comp.}\\
    \hline
    0$\%$ & 91.51 & 1 & 91.36 & 1 & 90.97 & 1 & 90.30 & 1 & 89.46 & 1 & 88.41 & 1  \\ \hline
    30$\%$  & 88.95 & 5.45 & 91.33 & 7.12 & 90.84 &  8.60 & 90.28 & 10.88 & 89.03 & 12.98 & 88.24 & 14.77  \\ \hline
    40$\%$ &  83.75 & 6.14 & 91.31 & 8.29 & 90.67 & 8.93 & 90.33 & 11.02 & 89.27 & 13.04 & 88.43 & 14.81   \\ \hline
    50$\%$ &  61.27 & 7.09 & 91.19 & 8.47 & 90.81 & 9.83 & 90.24 & 11.48 & 89.05 & 13.28 & 88.44 & 15.05  \\ \hline
    60$\%$ &  29.42 & 8.48 & 90.98 & 9.81 & 90.69 & 11.95 & 90.15 & 12.65 & 89.29 & 14.11 & 88.25 & 15.79  \\ \hline
    70$\%$ & 10.04 & 10.70 & 89.51 & 12.14 & 90.70 & 13.51 & 90.16 & 16.06 & 89.38 & 16.15 & 88.43 & 17.44  \\ \hline
    80$\%$ & 10.00 & 14.72 & 71.77 & 16.65 & 86.82 & 17.92 & 90.09 & 19.37 & 89.21 & 23.40 & 88.23 & 22.14  \\ \hline
    90$\%$ & 10.00 & 15.33 & 12.62 & 28.4 & 33.28 & 30.48 & 76.91 & 33.37 & 84.93 & 32.90 & 87.89 & 35.14   \\ \hline
  \end{tabular}
  \caption{Quantization Effect: Accuracy (Acc.) and Mask and Quantization Compression Ratio (Comp.) comparison to baseline at different sparsity levels.}
  \label{basecomp2}
\end{table*}

\begin{table*}
\centering
\begin{tabular}{ |c|c|c|c|c|c|c|  } 
 \hline
 \textbf{method} & \textbf{Original} & \textbf{CNET$_{\text{0.005}}$} & \textbf{CNET$_{\text{0.01}}$} & \textbf{CNET$_{\text{0.02}}$} & \textbf{CNET$_{\text{0.03}}$} & \textbf{CNET$_{\text{0.045}}$} \\ \hline
 \textbf{entropy} & 7.54 & 6.75 & 6.41 & 6.13 & 5.07 & 4.00 \\ \hline 

 \hline
\end{tabular}
	\caption{Entropy Experiments: Entropy at $70\%$ sparsity for different $\lambda$ settings.}
	\label{entropytable}
\end{table*}

For $k$-means clustering we always use 256 cluster centers to achieve 8-bit representations.
Since our method is trained with the compressibility loss, non-zero elements of the resulting $\bm{w_{net}}$ has low entropy, thus some clusters are more densely populated than others and samples are more compact around cluster centroids. 
Remember that in the ideal case of ending up in compressibility loss' local minima, $\bm{w_{net}}$ would have been ternary, thus the non-zero parts would have been binary. 
This would have resulted into two clusters whose elements are exactly equal to corresponding centroids and zero performance loss due to quantization.

In practice we never reach this case, however the low-entropy and compact-clusters effect is observed from the high gains in compression ratio compared to original model. In particular, when compressibility loss is weighted higher, i.e. when $\lambda$ is increased, at the same sparsity levels, the compression ratio shows a strong tendency to increase. For example at $90\%$ sparsity, the CNET$_{0.005}$ achieves 28.4 compression ratio and CNET$_{0.045}$ achieves 35.14 compression ratio, whereas the compression ratio obtained for the original model is only 15.33. This results from the compressibility aspect of our method for the non-zero elements. Remember here that the quantization is only applied to non-zero elements and at the same sparsity level,  CNET$_{0.005}$  and CNET$_{0.045}$ have same number of non-zero elements, thus the difference in compression ratio is directly due to compressibility of non-zero elements.  

In order to highlight the quantization effect even further, we also measure the entropy of the probability distribution function $P$ obtained by normalizing the vector of cluster populations. 
The entropy is simply measured via Eq. \ref{entropyeq}. It is clearly observed from Table \ref{entropytable} that as the weight on the loss is increased, entropy is decreased which is a direct indicator of increase in compressibility.

\begin{equation}
\label{entropyeq}
h=-\sum_i P_ilog_2(P_i)
\end{equation}

\begin{table}
\centering
\begin{tabular}{ |c|c|c|c|c|c|c|  } 
 \hline
 \textbf{Cl. No.} & 256 & 128 & 64 & 32 & 16\\ \hline
 \textbf{Acc.} & 87.89 & 88.23 & 84.61 & 81.65 & 45.4  \\ \hline 
 \textbf{Comp.} & 35.14 & 39.44 & 43.87 & 48.51 & 57.48 \\ \hline 

 \hline
\end{tabular}
	\caption{Cluster Number (Cl. No.) Experiments: accuracy (Acc.) and compression ratio (Comp.) for the method CNET$_{0.045}$ at $90\%$ sparsity}
	\label{cluster_exp}
\end{table}

Finally, we also perform experiments on the robustness to $k$-means centroid number of a particular method (CNET$_{0.045}$ at $90\%$ sparsity). This experiment also highlights the quantization effect of our method and enables achieving higher compression ratios with lower quantization centroid numbers. It can be observed from Table \ref{cluster_exp} that it is possible to achieve higher compression ratios without loss (128 clusters scenario) and moreover one can achieve $10\%$ lower accuracy than original model at around 50 compression ratio.

\subsection{On the Effect of Mask Compression}
We performed an experiment which does not apply mask compression but treats the zeros in the weights as a separate cluster with centroid 0. Then npz compression is applied to labels and to centroids. We performed the experiment for 50$\%$ sparsity for CNET$_{0.045}$ and achieved a compression ratio of 13.32, whereas with using mask compression one can achieve 15.05 compression ratio. This experiment justifies the use of mask compression and the compression method described in Section \ref{compmet}.

\subsection{Gradually Increasing $\lambda$ During Training}

In this experiment, we have gradually increased $\lambda$ during training.
In particular, we have started with $\lambda=0$ and linearly increased it by $0.007$ at the end of each epoch.
By this approach we could achieve very high compression ratios as one can see in Table \ref{ramp}.

\subsection{Comparison With the State-of-the-Art}

We compare our method with three recent state-of-the-art methods, namely Targeted Dropout (TDrop) \cite{Gomez2018}, Smallify \cite{Leclerc2018} and Universal DNN compression (UDNN) \cite{Choi2018}. 
While TDrop and Smallify are designed to achieve a high level of sparsity, they do not consider the quantization aspect of the non-zero weights. UDNN is mostly focused on the quantization aspect. Therefore we compare our method to Smallify and TDrop based on sparsity and to UDNN based on compression ratios.
For TDrop, $\alpha=0.66 , \gamma=0.75$ setting was reported which achieved the best top accuracy according to the original paper. The Smallify results were copied from \cite{Gomez2018}.
First, we report performance of CNET$_{0.045}$, Smallify and TDrop at different sparsity levels.

One can observe from Table \ref{soa1} that with fixed parameter settings, CNET and Smallify perform better than TDrop at highest sparsity level of $90\%$. Smallify method seems to outperform our method at the sparsity levels reported in table \ref{soa1}. However, our end goal is not to obtain a certain sparsity, but to achieve higher compression rates where sparsity is only one aspect. Another important note here is that the baseline performance of TDrop is reported to be 94.30, whereas ours and Smallify's baseline performance was around 91.50, therefore one should account for the 3 points of difference between the two baselines when comparing. 

\begin{table}
\centering
\begin{tabular}{ |c|c|c|c|c|c|c|  } 
 \hline
 \textbf{Sparsity} & 90$\%$ & 95$\%$ & 96$\%$ & 97$\%$ & 98$\%$\\ \hline
 \textbf{Acc.} & 86.44 & 85.79 & 86.19 & 85.12 & 76.57  \\ \hline 
 \textbf{Comp.} & 46.85 & 57.41 & 65.48 & 78.46 & 102.76 \\ \hline 

 \hline
\end{tabular}
	\caption{CNET$_{\text{ramp}}$ experiments with gradually increasing $\lambda$ during training: accuracy (Acc.) and compression ratio (Comp.) at indicated sparsity levels.}
	\label{ramp}
\end{table}

\begin{table}
\centering
\begin{tabular}{ |c|c|c|c|c|c|c|  } 
 \hline
 \textbf{Sparsity} & \textbf{30\%} & \textbf{50\%} & \textbf{70\%} & \textbf{80\%} & \textbf{90\%}\\ \hline
 \textbf{TDrop} & 93.89 & 93.84 & 93.84 & 92.31 & 46.57  \\ \hline 
 \textbf{Smallify} & 91.52 & 91.51 & 91.57 & 91.55 & 91.45 \\ \hline 
 \textbf{CNET$_\text{0.045}$} & 88.24 & 88.44 & 88.43 & 88.23 & 87.89 \\ \hline

 \hline
\end{tabular}
	\caption{TDrop, Smallify and CNET$_{0.045}$ comparison: accuracies at indicated sparsity levels.}
	\label{soa1}
\end{table}

The next comparison is at higher sparsity levels where TDrop and CNET at ramp setting was reported, i.e. during training parameters of both methods were gradually increased to slowly adapt to sparsification/compression. The results are shared in Table \ref{soa2}.

\begin{table}
\centering
\begin{tabular}{ |c|c|c|c|c|c|c|  } 
 \hline
 \textbf{Sparsity} & \textbf{90\%} & \textbf{95\%} & \textbf{96\%} & \textbf{97\%} & \textbf{98\%}\\ \hline
 \textbf{TDrop$_{\text{ramp}}$} & 88.70 & 88.73 & 88.74 & 88.67 & 88.70  \\ \hline 
 \textbf{Smallify} & 91.48 & 90.30 & 87.54 & 70.29 & 33.04 \\ \hline 
 \textbf{CNET$_{\text{ramp}}$} & 86.44 & 85.79 & 86.19 & 85.12 & 76.57 \\ \hline

 \hline
\end{tabular}
	\caption{TDrop$_{\text{ramp}}$, Smallify and CNET$_{\text{ramp}}$ comparison: accuracies at indicated sparsity levels.}
	\label{soa2}
\end{table}

One can observe from Table \ref{soa2} that TDrop and CNET with ramp setting, can achieve very robust performances at very high sparsity levels whereas Smallify fails to do so. It is observed that at ramp settings TDrop is more robust than CNET at extremely high sparsity level $98\%$. We speculate that this might be the case due to inherent fine-tuning in TDrop method, where the weights are actually pruned during training, whereas in our method this is not the case. We aim to include improvements in this aspect in our future work.

\begin{figure}

\includegraphics[width=\linewidth]{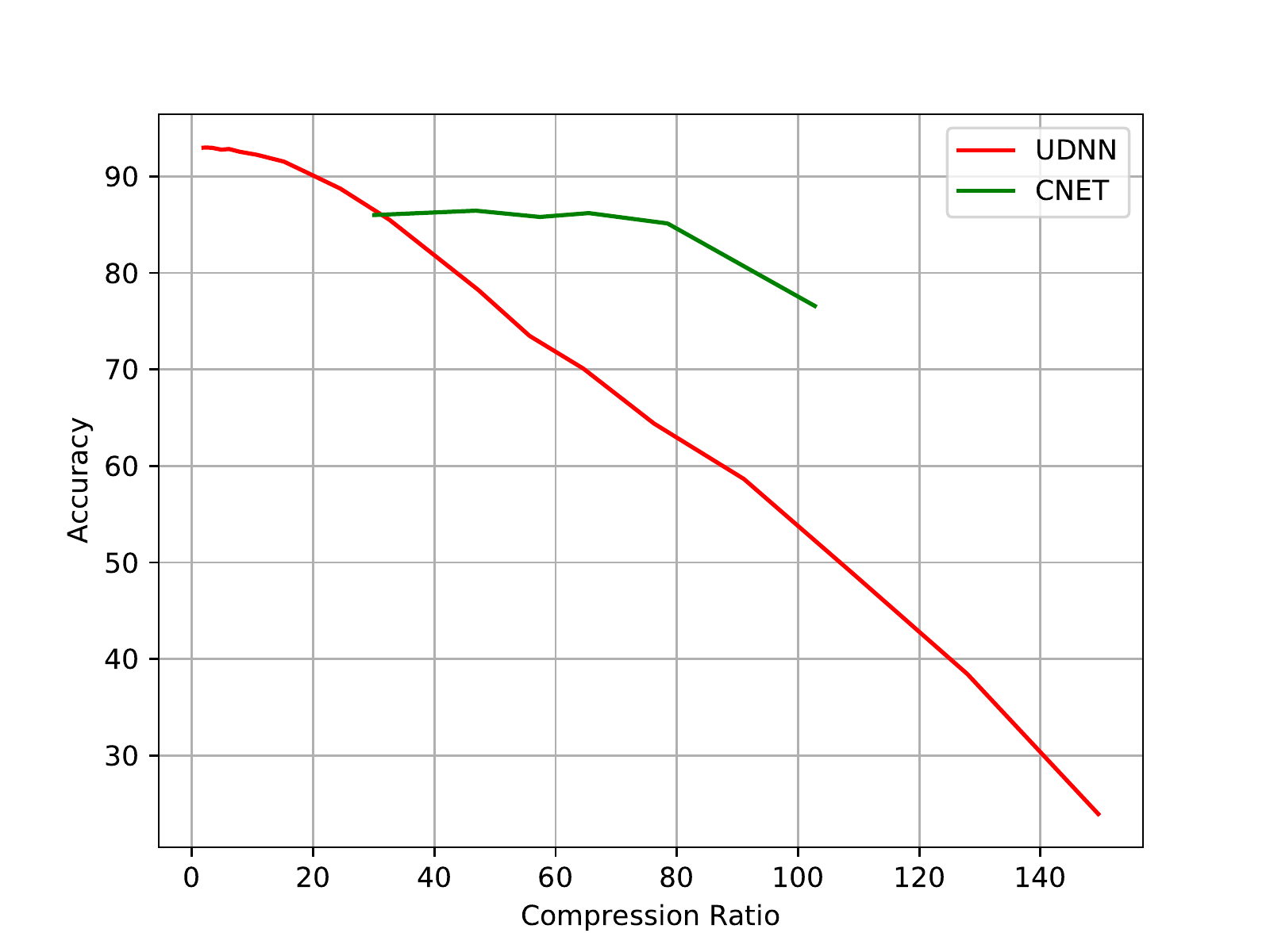}
\caption{Comparison with UDNN}
\label{fig:UDNNcomp}

\end{figure}

Since TDrop and Smallify do not make any experiments on the compression ratio, we also compare our method (CNET$_{\text{ramp}}$) to UDNN where compression ratio was reported. The experimental results for UDNN were provided by the authors of the paper. We choose the variants of both methods that achieve the highest compression rates. The results are provided in Fig. \ref{fig:UDNNcomp}. One can observe that at the given range of compression ratio, our method outperforms UDNN in terms of accuracy by a large margin.

\section{Conclusions}
\label{conclusions}
We have adopted a previously proposed negated sparsity measure and both theoretically and experimentally showed that when this measure is used as a compressibility loss, it results into more compressible neural network weights both in terms of sparsity and low-entropy non-zero part of the weights. We have also shown that our method is comparable or better to state-of-the-art methods on neural network compression.

\bibliographystyle{named}
\bibliography{ijcai19}

\appendix

\section{Proof of Theorem \ref{local}}\label{proofsec}

\begin{proof}
Let $c$ be a positive value and $\epsilon$ be a very small positive number satisfying $c>>\epsilon$. 
Let $\bm{\epsilon}$ be a vector where each element is $\epsilon_j \in \{-\epsilon, 0 ,\epsilon\}$.
At local minima $\bm{x}^*$ , for any $\bm{\epsilon}$, the following must hold:

\begin{equation}
L(\bm{x}^*+\bm{\epsilon})>L(\bm{x}^*)
\label{lm_criterion}
\end{equation}

We know from Theorem \ref{main} that at any critical point $\bm{x}$, $x_i \in \{ -c,0,c \}$.
Let $N$ be the set of non-zero elements of $\bm{x}$ and $Z$ be the set of zero elements of $\bm{x}$.
Let us start by writing the following expansion:



We know from Corollary \ref{maincorol} that at any critical point $\bm{x}$, $L(\bm{x})=\sqrt{n}$.
Then, one can rewrite Eq. \ref{lm_criterion} as follows:

\begin{equation}
    \frac{\sum_{i \in Z} |\epsilon_i| + \sum_{j \in N} |x_j +\epsilon_j|}{\sqrt{\sum_{i \in Z} \epsilon_i^2 + \sum_{j \in N} (x_j +\epsilon_j)^2}}>\sqrt{n}
    \label{Eq14}
\end{equation}

By taking square of both sides of the inequality and rearranging one can rewrite Eq. \ref{Eq14} as follows.



\begin{equation}
(\sum_{i \in Z} |\epsilon_i| + \sum_{j \in N} |x_j +\epsilon_j|)^2>n \sum_{i \in Z} |\epsilon_i|^2 +  n \sum_{j \in N} (x_j +\epsilon_j)^2
\label{Eq15}
\end{equation}

\begin{multline}
(\sum_{i \in Z} |\epsilon_i|)^2 + (\sum_{j \in N} |x_j +\epsilon_j|)^2 + 2 \sum_{i \in Z} \sum_{j \in N} |\epsilon_i| |x_j +\epsilon_j|  >  \\
n \sum_{i \in Z} |\epsilon_i|^2 +  n \sum_{j \in N} (x_j +\epsilon_j)^2
\label{Eq16}
\end{multline}

\begin{lemma}
\label{lemma21}
The following inequality holds.

\begin{equation}
(\sum_{i \in Z} |\epsilon_i|)^2+ 2 \sum_{i \in Z} \sum_{j \in N} |\epsilon_i| |x_j +\epsilon_j|  - n \sum_{i \in Z} |\epsilon_i|^2  \geq 0 
\label{Eq17}
\end{equation}

\end{lemma}
\begin{proof}

One can rewrite Eq. \ref{Eq17} as follows.

\begin{equation}
(z_n \epsilon)^2+ 2  z_n \epsilon \sum_{j \in N}  |x_j +\epsilon_j|  - n z_n \epsilon^2 \geq 0
\label{Eq18}
\end{equation}

Note that $z_n$ is the total number of non-zeros in $\bm{\epsilon}$ vector in the region where $x_i=0$. 
Then, one can deduct that

\begin{equation}
 \sum_{j \in N}  |x_j +\epsilon_j|= \sum_{j \in N} c + \sum_{k \in N_+} \epsilon - \sum_{k \in N_-} \epsilon
 \label{Eq19}
\end{equation}

\begin{equation}
 \sum_{j \in N}  |x_j +\epsilon_j|=  nc + a \epsilon
\label{Eq20}
\end{equation}

In Eq. \ref{Eq20},  $a=|N_+|-|N_-|$ where $N_+$ is the set where $sign(x_j)=sign(\epsilon_j)$ holds and  $N_-$ is the set where $sign(x_j)=-sign(\epsilon_j)$ holds. Putting Eq. \ref{Eq20} in place in Eq. \ref{Eq18}, one can rewrite as:

\begin{equation}
(z_n \epsilon)^2+ 2  z_n \epsilon nc + 2 z_n \epsilon a \epsilon  - n z_n \epsilon^2 >=0 
\label{Eq21}
\end{equation}

Which reduces to:

\begin{equation}
(z_n \epsilon) (z_n \epsilon +2nc + 2a \epsilon -n \epsilon) >=0
\label{Eq22}
\end{equation}

Notice that since $c>>\epsilon$, $2nc>\epsilon(2a+z_n-n)$ holds.
This is easy to see since it is equivalent to $c>\epsilon\frac{(2a+z_n-n)}{2n}$ and  $\frac{(2a+z_n-n)}{2n}< \frac{(2z-n)}{2n}$ hence $\frac{(2a+z_n-n)}{2n}< 0.5$.
Therefore it is straightforward to see that $c>0.5 \epsilon$

This reduces Eq. \ref{Eq22} to : $z_n>=0$.
Hence Lemma \ref{lemma21} is proven.

\end{proof}

\begin{lemma}
\label{lemma22}
The following inequality holds.

\begin{equation}
(\sum_{j \in N} |x_j +\epsilon_j|)^2  < n \sum_{j \in N} (x_j +\epsilon_j)^2
\label{Eq25}
\end{equation}

\end{lemma}

\begin{proof}
From Eq. \ref{Eq20}, one can rewrite Eq. \ref{Eq25} as follows.

\begin{equation}
(nc+a\epsilon)^2  < n \sum_{j \in N} (x_j^2 +\epsilon_j^2 + 2x_j \epsilon_j)
\label{Eq26}
\end{equation}

One can rewrite Eq. \ref{Eq26} as follows.

\begin{equation}
(nc)^2+(a\epsilon)^2+ 2nca\epsilon  < n^2c^2 + n n_n \epsilon^2 + n \sum_{j \in N} 2x_j \epsilon_j
\label{Eq27}
\end{equation}

Note that $n_n$ is the total number of non-zeros in $\bm{\epsilon}$ vector in the region where $x_i \neq 0$. 


Note that one can write:


\begin{equation}
n \sum_{j \in N} 2x_j \epsilon_j =  2nca\epsilon
\label{Eq30}
\end{equation}

Putting Eq. \ref{Eq30} in place in Eq. \ref{Eq27} gives the following.



\begin{equation}
\epsilon^2((a)^2 - n n_n) \leq 0 
\label{Eq32}
\end{equation}

Note that maximum value $a$ can take is $n_n$, therefore Eq. \ref{Eq32} is always satisfied since $n_n \leq n$ holds.
Hence Lemma \ref{lemma22} is proven.

\end{proof}

Note that Lemma \ref{lemma21} and \ref{lemma22} presents opposite aspects for the inequality in Eq. \ref{Eq16}.
Hence one should combine the intermediate results Eqs. \ref{Eq22} and \ref{Eq32} and rewrite Eq. \ref{Eq16} as follows.

\begin{equation}
(z_n ) (z_n \epsilon +2nc + 2a \epsilon -n \epsilon) > \epsilon^2 (nn_n-a^2) 
\label{Eq33}
\end{equation}

Eq. \ref{Eq33} can reduce to the following by dividing each side with $\epsilon$.

\begin{equation}
(z_n ) (z_n +\frac{2nc}{\epsilon } + 2a  -n ) > \epsilon (nn_n-a^2) 
\label{Eq34}
\end{equation}

Note that the right side of the equation is in fact a very small value due to multiplication by $\epsilon$. Also notice the very large number $\frac{2nc}{\epsilon }$ in the left side.
Therefore as long as $z_n \neq 0$ is satisfied, Eq. \ref{Eq34} holds, thus proving the Theorem \ref{local}.
\end{proof}

\end{sloppypar}
\end{document}